\documentclass[letterpaper]{article} 
\usepackage{aaai2026}  
\usepackage{times}  
\usepackage{helvet}  
\usepackage{courier}  
\usepackage[hyphens]{url}  
\usepackage{graphicx} 
\usepackage{natbib}  
\usepackage{caption} 
\usepackage{algorithm}
\usepackage{algorithmic}
\usepackage{color}
\usepackage{newfloat}
\usepackage{listings}
\usepackage{multirow}
\usepackage{booktabs}
\usepackage{amsmath}
\usepackage{xcolor} 
\usepackage{colortbl} 
\usepackage{amsfonts}
\usepackage{subcaption}
\definecolor{macaroongreen}{RGB}{255, 220, 220} 
\usepackage{times}
\usepackage{helvet}
\usepackage{courier}
\usepackage{newfloat}
\usepackage{listings}
\usepackage{newtxtext}
\usepackage{newtxmath}
\DeclareCaptionStyle{ruled}{labelfont=normalfont,labelsep=colon,strut=off} 
\floatstyle{ruled}
\newfloat{listing}{tb}{lst}{}
\floatname{listing}{Listing}
\title{Continuous Vision-Language-Action Co-Learning with 
Semantic-Physical Alignment for Behavioral Cloning}
\author{
    Xiuxiu Qi\textsuperscript{\rm 1, 2}, Yu Yang\textsuperscript{\rm 3}, Jiannong Cao\textsuperscript{\rm 2}\thanks{Corresponding Authors.}, Luyao Bai\textsuperscript{\rm 2}, \\
    Chongshan Fan\textsuperscript{\rm 1}, 
    Chengtai Cao\textsuperscript{\rm 4}, Hongpeng Wang\textsuperscript{\rm 1}\footnotemark[1]
}
\affiliations{
    \textsuperscript{\rm 1}The College of Artificial Intelligence \& Shenzhen Research Institute, Nankai University, Tianjin, China.\\
    \textsuperscript{\rm 2}Department of Computing, The Hong Kong Polytechnic University, Hong Kong SAR, China.\\
    \textsuperscript{\rm 3}Centre for Learning, Teaching and Technology, The Education University of Hong Kong, Hong Kong SAR, China.\\
    \textsuperscript{\rm 4}Department of Computer Science, City University of Hong Kong, Hong Kong SAR, China.\\

}

\usepackage{bibentry}

\begin{document}

\maketitle

\begin{abstract}
Language-Conditioned Manipulation (LCM) facilitates human-robot interaction via Behavioral Cloning (BC), which learns control policies from human demonstrations and serves as a cornerstone of embodied AI. Overcoming compounding errors in sequential action decisions remains a central challenge to improving BC performance. Existing approaches mitigate compounding errors through data augmentation, expressive representation, or temporal abstraction. However, they suffer from physical discontinuities and semantic-physical misalignment, leading to inaccurate action cloning and intermittent execution. In this paper, we present Continuous vision-language-action Co-Learning with Semantic-Physical Alignment (CCoL), a novel BC framework that ensures temporally consistent execution and fine-grained semantic grounding. It generates robust and smooth action execution trajectories through continuous co-learning across vision, language, and proprioceptive inputs (i.e., robot internal states). Meanwhile, we anchor language semantics to visuomotor representations by a bidirectional cross-attention to learn contextual information for action generation, successfully overcoming the problem of semantic-physical misalignment. Extensive experiments show that CCoL achieves an average 8.0\% relative improvement across three simulation suites, with up to 19.2\% relative gain in human-demonstrated bimanual insertion tasks. Real-world tests on a 7-DoF robot further confirm CCoL’s generalization under unseen and noisy object states.
\end{abstract}

\section{Introduction}
\label{introduction}
Language-Conditioned Manipulation (LCM) is an emerging field of research in embodied AI and a stepping stone toward artificial general intelligence \cite{stepputtis2020language,
karamcheti2023language,
pan2025self}. This domain seeks to develop an embodied agent to understand its surrounding environments and perform complex, goal-directed behaviors
based on human instructions.
As an imitation learning paradigm, Behavioral Cloning (BC) bridges high-level task instructions with low-level robotic control \cite{li2024learning,kelly2019hg,wang2024explore}. Traditional BC methods 
mimic expert demonstrations through state-action mappings, enabling significant robotic applications, from self-driving vehicles \cite{codevilla2019exploring} to robot manipulation \cite{cui2021toward}. For LCM tasks,
BC extends this paradigm by incorporating language as an additional modality, allowing robots to generalize executable actions via multimodal inputs. This integration is particularly critical in unstructured environments where tasks demand precise alignment between semantic intent (e.g., object affordances) and physical execution (e.g., end-effector trajectories).

Despite its promise, BC suffers from compounding errors caused by sequential error propagation, resulting in a covariate shift \cite{ingram2023creating,mehta2025stable}.
Semantic-physical misalignment between multimodal inputs amplifies these errors (e.g., misaligned object affordances or trajectory deviations), leading to catastrophic failures in long-horizon LCM tasks as even minor inaccuracies accumulate quadratically over time \cite{block2024provable}.
To address this challenge, prior works have explored three primary directions: (1) 
data augmentation techniques such as noise injection and synthetic data generation \cite{
deshpande2024data}, which enhance training diversity and improve error approximation, and interactive correction methods that iteratively collect expert feedback during rollout \cite{hoque2023fleet};
 (2) expressive representation spaces that aim to minimize single-step prediction errors by leveraging semantically fused multimodal features \cite{ke20243d}; (3) temporal abstraction methods that reduce decision-making horizon by treating long action sequences as single, high-level primitives \cite{
 shiwaypoint,
 lukoi}.

However, existing methods still face two key multimodal grounding challenges impacting action cloning and execution coherence. First, physical discontinuities arise from discrete action modeling paradigms, such as temporal abstraction \cite{fu2024mobile}, which disrupt essential motion continuity.
For instance, abrupt dual-arm waypoint transitions during peg insertion tasks result in jerky trajectories with non-smooth acceleration profiles, ultimately causing failures in long-horizon tasks due to incoherent action execution. 
Second, semantic-physical misalignment occurs when high-level semantic goals
fail to guide physical actions accurately. Static fusion methods like R3M \cite{nair2023r3m}
globally align language and vision but overlook stepwise semantic adaptation, resulting in inaccurate action cloning. For example, executing 
``place the cup on the shelf'' requires the robot to dynamically shift attention from the cup during grasping to the shelf during placement.

To address these challenges, we present a novel BC framework, \textbf{\underline{C}}ontinuous Vision-Language-Action \textbf{\underline{Co}}-\textbf{\underline{L}}earning with Semantic-Physical Alignment (\textbf{CCoL})\footnote{Project website is available at https://qhemu.github.io/CCoL/.
}. 
CCoL introduces continuous co-learning across vision, language, and proprioceptive inputs to ensure temporally consistent latent trajectories, while stepwise anchoring language semantics to visuomotor representations to enable semantic-to-physical correspondence. 
Specifically, the multimodal continuous co-learning mechanism captures proprioceptive dynamics in latent space by Neural Ordinary Differential Equations (NeuralODEs) \cite{bilovs2021neural,li2024hides} and enables joint fusion across vision, language, and proprioceptive inputs. 
Meanwhile, we design a cross-modal semantic-physical alignment mechanism that considers language as a flexible target specification and progressively anchors language semantics to visuomotor representations via a bidirectional cross-attention mechanism.
Finally, a goal-conditioned decoder generates action sequences from enriched fusion representations, with a hybrid loss jointly optimizing action accuracy and latent state smoothness for stable training and physically feasible outputs. Our contributions are as follows:
\begin{itemize}
    \item We propose a multimodal continuous co-learning mechanism integrating vision, language, and proprioception to model temporal dependencies in latent space, ensuring smooth and consistent transitions across action states.
    \item We introduce a cross-modal semantic-physical alignment module that anchors linguistic specifications to visuomotor representations at each step, enabling fine-grained semantic adaptation during task execution.
    \item We conduct extensive experiments on three simulation suites, showing that CCoL achieves SOTA performance and proves real-world effectiveness with a 7-DoF robot.
\end{itemize}

\section{Preliminaries}
\label{preliminaries}


 \begin{figure*}[!t]
	\centering
 \includegraphics[width=0.956\hsize]
 {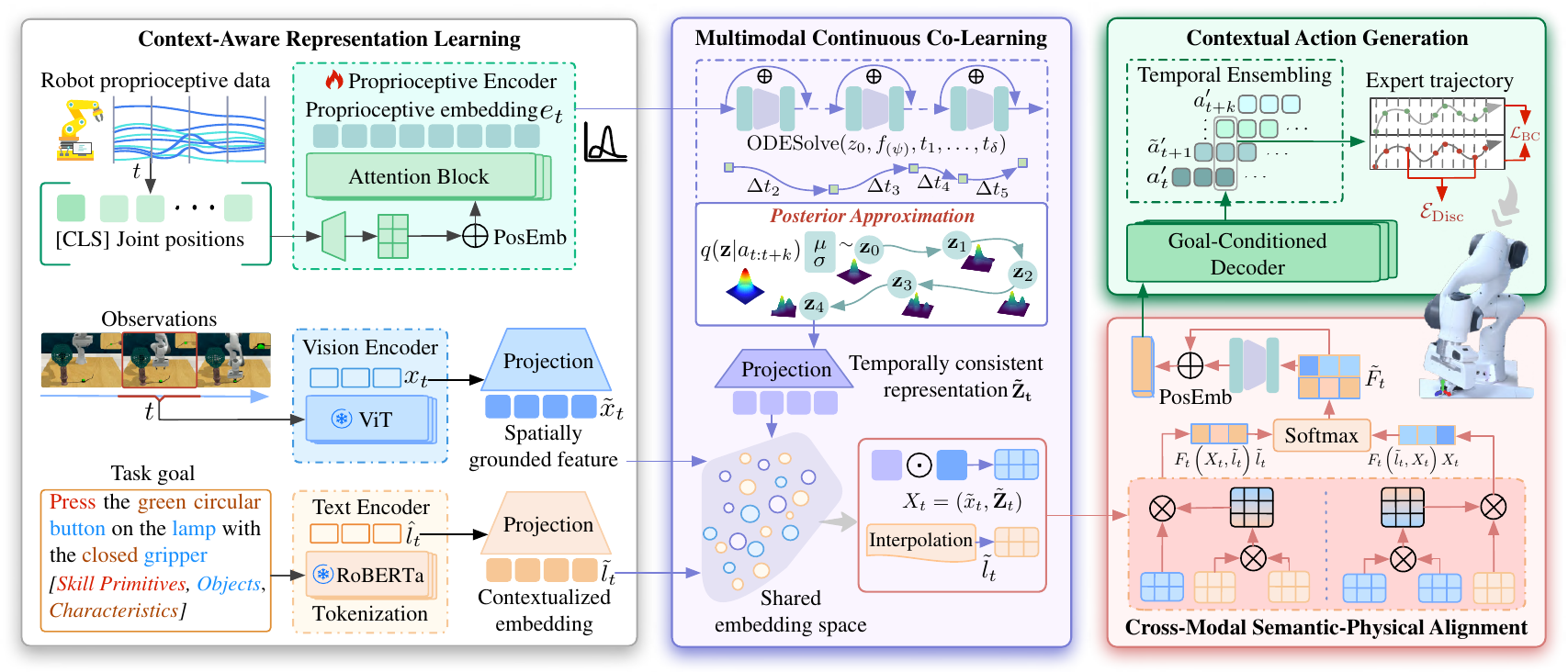}
	\caption{
    Overview of the CCoL framework. 
    MCC leverages dynamic proprioceptive modeling to capture temporal evolution and maps vision, language, and proprioception into a shared latent space (\textbf{purple frame}). 
    CSA synchronizes stepwise semantic information across modalities (\textbf{red frame}) to enable the generation of contextually 
    and physically feasible action sequences.
	}
	\label{opmodel}
 \end{figure*} 
 
LCM leverages BC methods to infer executable behaviors from heterogeneous inputs, enabling precise and generalizable motor control~\cite{
stepputtis2020language,
karamcheti2023language,
livision}.
Formally, we assume access to a dataset $\mathcal{D} = \{l, \tau_i\}_{i=1}^N$ of $N$ expert demonstrations, where each demonstration has a sequence of state-action pairs over horizon $H$, $\tau = \{(s_1,a_1),(s_2,a_2),\dots,(s_H,a_H)\}$, aligned with a task goal $l \in L$ (e.g., ``slide the red cube''). The states $s_t=(o_t, r_t) \in S$ integrate current visual observation $o_t$ and robot proprioceptive data $r_t$, while actions $a_t \in A$ represent motor commands like end-effector positions or joint velocities.
Our goal is to learn a policy $\pi_{\theta}: S \times L \rightarrow A$, which maps states $S$ and natural language instructions $L$ to actions $A$. We optimize 
this policy $\pi_{\theta}$ by minimizing the negative log-likelihood of the observed expert actions:
\begin{equation}
    \mathcal{L}_{\pi_{\theta}} =\underset{(\tau_i, l) \sim \mathcal{D}}{\mathbb{E}}\left[\sum_{t=1}^{H} -\log \pi_\theta(a_{t}|s_{t}, l)\right].
\end{equation}
While LCM extends traditional BC with language guidance, it is still subject to critical \textit{compound error} limitations.

\subsubsection{Single-Step Error Propagation.} Mismatched vision-action mappings induce stepwise prediction errors,
causing a covariate shift.
Let ${a}_{t}^{\prime}=\pi_{\theta}(s_{t},l)$ denote predicted action. Stepwise error $\epsilon = ||{a}_t^{\prime} - a_t||^2$ propagates through the transition dynamics. 
Since transitions are sequential, 
the cumulative error grows quadratically with $H$ rather than linearly: $H\epsilon+(H-1)\epsilon+ \dots +\epsilon \sim O(H^{2}\epsilon)$ \cite{xu2020error}. 
\subsubsection{Discontinuous Action Sequences.} 
Discretized action predictions (e.g., time abstraction) induce \textit{higher-order physical infeasibility} through piecewise-constant control signals. These abrupt transitions violate differential continuity constraints (e.g., non-smooth acceleration profiles, discontinuous contact forces \cite{righetti2013optimal}).
Our key observation is that physical feasibility requires \textit{continuous temporal coherence} beyond adjacent actions. We  propose to regularize the action trajectory through dynamical consistency:
\begin{equation}
\resizebox{0.54\columnwidth}{!}{
$\mathcal{E}_{\mathrm{disc}} = \int_{t}^{t+\Delta t} \rho\left(\frac{da}{dt}, \frac{d^2a}{dt^2}\right) dt,$}
\end{equation}
where $\rho(\cdot)$ encodes physical constraints. To address both single-step inaccuracies and temporal discontinuity, we propose a unified optimization framework:
\begin{align}
\resizebox{0.9\columnwidth}{!}{
$\mathcal{L}_{\pi_{\theta}}=\mathbb{E}_{(\tau_i, l) \sim \mathcal{D}}\sum_{t=1}^{H}\left(\underbrace{\|\pi_\theta\left(s_{t}, l\right)-a_t\|_2^2}_{\text{Stepwise Loss}}
    +\underbrace{
    \mathcal{E}_{\mathrm{disc}}}_{\text {Disc Penalty}}\right),
    $
    }
\end{align}
where the stepwise loss ensures local alignment between learned and expert actions, and the discontinuity penalty regularizes action sequences to ensure physical feasibility.

\section{Methodology}
\label{methodology}

As depicted in Fig. \ref{opmodel}, CCoL primarily consists of two key components: \textit{Multimodal Continuous Co-Learning} (MCC) and \textit{Cross-Modal Semantic-Physical Alignment} (CSA).

\subsection{Context-Aware Representation Learning}
We first introduce context-aware representation learning to ensure that each modality is independently encoded while preserving its unique characteristics. 

\noindent\textbf{Vision Encoder.} 
The vision encoder processes RGB-D video frames \(o_t \in \mathbb{R}^{H \times W \times 3}\) using a 
Vision Transformer (ViT) \cite{han2022survey}. The encoder extracts spatially grounded features \(x_t = \operatorname{ViT}(o_t), x_t \in \mathbb{R}^{d_v}\), capturing spatial cues critical for grounding language in the visual scene.

\noindent\textbf{Text Encoder.} 
To process language instructions, the text encoder employs a RoBERTa model \cite{liu2019roberta}. Instructional texts are tokenized
and encoded 
into contextualized embeddings: \(\hat{l}_t = \operatorname{RoBERTa}(l)\), where \(\hat{l}_t \in \mathbb{R}^{d_l}\).

\noindent\textbf{Proprioceptive Encoder.} 
The proprioceptive encoder processes robotic internal states, including position sequences \(r_t \in \mathbb{R}^{k}\).
Inspired by \cite{zhaolearning}, we implement a Conditional Variational Autoencoder (CVAE) to map these inputs into a latent motion pattern space. 
The inputs $r_t$ are first processed through linear transformations and concatenated with a [CLS] token. Sinusoidal positional encodings are added to incorporate sequential order information. The concatenated sequence is then processed by a Transformer \cite{vaswani2017attention} or a Temporal Convolutional Network (TCN) \cite{lea2017temporal}, producing the proprioceptive embedding \(e_t = \operatorname{CVAE}([\operatorname{CLS}; r_t])\), where \(e_t \in \mathbb{R}^{d_a}\).

\subsection{Multimodal Continuous Co-Learning}
Conventional BC approaches often suffer from temporally fragmented action sequences, where decoupled per-step predictions fail to account for underlying motion dynamics. Such limitations lead to jerky trajectories (e.g., high-jerk robotic arm movements) or kinematically invalid transitions (e.g., sudden end-effector orientation flips). 
Inspired by the notion of co-learning from \cite{liu2024multi}, we introduce a novel \textit{Multimodal Continuous Co-learning} (MCC) that uses NeuralODEs to capture the continuous evolution of proprioceptive embeddings. By integrating multimodal embeddings, 
we construct a shared latent space that captures temporal coherence and task-specific adaptability.
Specifically, we project proprioceptive 
embedding
at [CLS] token to predict the parameters of diagonal Gaussian distribution:
\begin{align}\nonumber
    \boldsymbol{\mu} &= {W}_{\mu} {e}_{t}[\text{CLS}] + {b}_{\mu}, \quad \\
    \log \boldsymbol{\sigma}^2& = {W}_{\sigma}{e}_{t}[\text{CLS}] + {b}_{\sigma},
\end{align}
where ${W}_{\mu}, {W}_{\sigma}$ and ${b}_{\mu}, {b}_{\sigma}$ are learnable parameters.
The latent state $z_0$ is initialized via
a conditional variational encoder with the reparameterization trick:
\begin{equation}
z_0 = \boldsymbol{\mu} + \boldsymbol{\varepsilon} \cdot \exp\left(\frac{1}{2} \log \boldsymbol{\sigma}^2\right), \quad \boldsymbol{\varepsilon} \sim \mathcal{N}(0, 1),  
\end{equation}
where \( \boldsymbol{\varepsilon} \) is a noise vector sampled from a standard normal distribution \( \mathcal{N}(0, 1)\). 
The NeuralODEs models the continuous evolution of this latent state as the solution to an initial value problem defined by a differential equation:
\begin{equation}
	{z}(t_{\delta}) = {z}_0 + \int_{0}^{t_{\delta} } f({z}(t), t; \psi) \, dt,
    \label{eqode}
\end{equation}
where $t_{\delta}$ represents the elapsed integration time, and $f({z}(t), t; \psi)$ is a residual MLP that learns the derivatives of the latent state with respect to time. 
Parameters $\psi$ are trained to ensure that latent trajectories match the underlying proprioceptive dynamics.
We solve this differential equation numerically using adaptive-step Dormand-Prince solvers via \texttt{odeint}, which computes $\mathbf{Z_t} = \texttt{odeint}(f, {z}_0, t)$ as the latent states over discrete time points $t$.
These latent trajectories replace stepwise proprioceptive features \({e}_t\), providing temporally consistent representations that mitigate fragmentation and discontinuities in conventional encoders.

To enable joint reasoning across modalities, visual features \(x_t\), language features \(\hat{l}_t\), and proprioceptive features $\mathbf{Z_t}$ are projected into a shared embedding space:
\begin{align}\nonumber
    \tilde{x}_t &= \operatorname{ReLU}(W_v x_t + b_v), \\
    \quad \tilde{l}_t &= \operatorname{ReLU}(W_l \hat{l}_t + b_l), \\\nonumber
    \quad \mathbf{\tilde{Z}_t} &= \operatorname{ReLU}(W_a \mathbf{Z}_t + b_a).
\end{align}
Here, \(W_v \in \mathbb{R}^{d_v \times h}\), \(W_l \in \mathbb{R}^{d_l \times h}\) and \(W_a \in \mathbb{R}^{d_a \times h}\) map original features to a higher dimension \(h\), while \(b_v, b_l, b_a \in \mathbb{R}^{h}\) are learnable biases.  
Moreover, language embeddings are upscaled to match the resolution of visual features via bilinear interpolation, enabling pixel-wise synchronization for fine-grained cross-modal fusion. 
\begin{figure}[t]
	\centering
	\includegraphics[width=0.985\hsize]
    {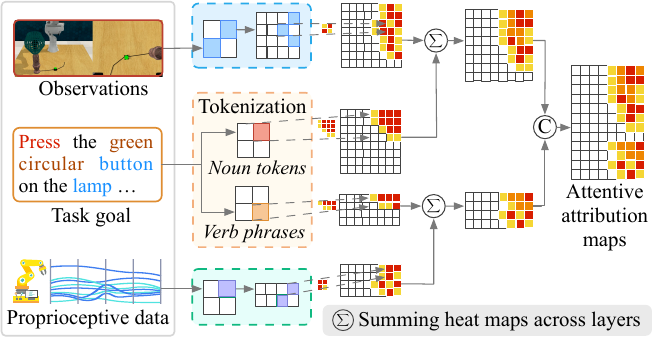}
	\caption{Illustration of attentive attribution maps
    (e.g., noun grounding 
    and verb-conditioned trajectory adjustment). 
    }	\label{attmaps2}
\end{figure}

\subsection{Cross-Modal Semantic-Physical Alignment}
Traditional feature fusion methods fail to dynamically align linguistic instructions with visual observations and proprioceptive states, causing inconsistencies in cross-modal grounding. To this end,
we design a \textit{Cross-Modal Semantic-Physical Alignment} (CSA) that anchors linguistic concepts to visuomotor representations at each timestep. As shown in Fig. \ref{attmaps2}, the normalized attention scores directly decode linguistic grounding
into physically constrained pixel/trajectory attributions,
ensuring semantic-to-physical correspondence. 
Specifically, we design an attentive attribution mapper that anchors lexical semantics to visuomotor representations at each timestep via bidirectional cross-attention.
Formally, at each attention head $\iota \in [1, n]$, given text embedding $\tilde{l}_{t}$ and joint vision-proprioceptive context $X_t=(\tilde{x}_t, \tilde{\mathbf{Z}}_t)$, the attention score determines the token-wise relevance:
\begin{align}\nonumber
F(\iota)_t\left(\tilde{l}_{t}, X_t\right)=\operatorname{softmax}\left(\frac{\left(W_q^{(\iota)} \tilde{l}_{t}\right)\left(W_k^{(\iota)} X_t\right)^T}{\sqrt{d_k}}\right),\\
F(\iota)_t\left(X_{t},\tilde{l}_t\right)=\operatorname{softmax}\left(\frac{\left(W_q^{(\iota)} X_{t}\right)\left(W_k^{(\iota)} \tilde{l}_{t}\right)^T}{\sqrt{d_k}}\right),
\end{align}
where $W_q^{(\iota)}, W_k^{(\iota)} \in \mathbb{R}^{h \times d_k}$ 
are learned projection matrices for queries and keys, $d_k$ is the dimension of the key vectors. These attention scores determine the alignment between linguistic tokens (e.g., verbs and nouns) and physical features, ensuring semantic-to-physical correspondence. The fused features \(\tilde{F}_t\) are calculated as follows:
\begin{equation}
    \tilde{F}_t=\sum_{n} F\left(\tilde{l}_{t}, X_t\right) X_{t,n} + F\left(X_{t},\tilde{l}_t\right) \tilde{l}_{t,n},
\end{equation}
where each term is broadcast across the sequence length and feature dimensions.
To maintain temporal coherence, we incorporate positional encodings $\mathbf{pos}\in \mathbb{R}^h$ into the self-attention mechanism. 
Specifically, the queries and keys are generated from these encodings through learned projections $W_q^{(\iota)}$ and $W_k^{(\iota)}$ to capture position-dependent relationships:
\begin{equation}
    \xi_t=\sum_n \operatorname{softmax}\left(\frac{\left(W_q^{(\iota)} \mathbf{pos}_{t}\right)\left(W_k^{(\iota)} \mathbf{pos}_{t}\right)^T}{\sqrt{d_k}}\right)\tilde{F}_{t,n}.
\end{equation}
Here, $\xi_t$ represents the final multimodal representation at 
$t$, incorporating both semantic alignment and temporal coherence. 
CSA overcomes the limitations of static fusion methods by grounding linguistic instructions in visuomotor representations via bidirectional cross-attention. Stepwise semantic anchoring aligns linguistic tokens (e.g., ``button'', ``press'') with visual regions and proprioceptive states, while positional encodings ensure temporal coherence.

\subsection{Contextual Action Generation}
The goal-conditioned decoder predicts future joint positions over the next \(k\) timesteps, ensuring contextual relevance and robustness to distributional shifts. Residual connections, layer normalization, and dropout are used to stabilize training and improve output quality.
\begin{equation}
    {a}^{\prime}_{t:t+k}=\operatorname{LayerNorm}(\tilde{F}_{t}+\operatorname{Dropout}(\xi_t)).
\end{equation} 
Each attention block is followed by a position-wise feed-forward network, and the processed features (dimension \(k \times d_k\)) are down-projected through an MLP into \(k \times d\), representing the predicted action sequences for the next $k$ timesteps. 
During training, the policy $\pi_{\theta}$ is optimized via supervised learning on expert demonstrations, minimizing the discrepancy between predicted and expert actions.
During inference, only the decoder is active, ensuring deterministic outputs for consistent evaluation.
The decoder generates action sequences $a^{\prime}$ by conditioning on real-time images and robot 
state $s^{\prime}$ and task instructions $l^{\prime}$, while it is executed under the environment dynamics (e.g., variations in position).

\subsection{Posterior Approximation and Optimization}
To model the complex posterior distribution of the latent variable \(z\), we follow the CVAE framework. The true posterior \(p_{\psi}(z|{a}_{t:t+k})\) is approximated by \(q_{\phi}(z|{a}_{t:t+k})\), parameterized by a neural network. The optimization objective is to maximize the Evidence Lower Bound (ELBO), comprising reconstruction and KL divergence losses:
\begin{equation}
\begin{aligned}
\log p({a}^{\prime}_{t:t+k})  \geq&   \underbrace{\mathbb{E}_{q(z|{a}^{\prime}_{t:t+k})}[\log p({a}^{\prime}_{t:t+k}|z,s,l)]}_{\text {Reconstruction Likelihood}}\\
& -\underbrace{\operatorname{KL}(q(z|{a}^{\prime}_{t:t+k}) \| p(z))}_{\text {Kullback-Leibler divergence }},
\end{aligned}
\end{equation}
where $p({a}^{\prime}_{t:t+k}|z,s,l)$ is the likelihood of the data given the latent variables, and $p(z)$ is the prior distribution of the latent variables. By maximizing the ELBO, we effectively maximize the log probability of the data while ensuring that the variational distribution $q(z|{a}^{\prime}_{t:t+k})$ is close to the true posterior.
The reconstruction loss ensures that decoded trajectories align with expert demonstrations, while the KL divergence regularizes the latent encoding to conform to a standard Gaussian prior, $\mathcal{L}_{\mathrm{BC}} = \mathcal{L}_{\mathrm{recon}} + \mathcal{L}_{\mathrm{KL}}$.
However, the estimated posterior $q_{\phi}(z|{a}^{\prime}_{t:t+k})$ in vanilla CVAE is restricted to a normal distribution, which limits its ability to reflect the real trajectory characteristics.
By employing NeuralODEs, we enhance the posterior's adaptability by learning dynamic representations of latent trajectories, better reflecting real-world data dynamics. Additionally, a discontinuity penalty is introduced to ensure the smooth evolution of latent states:
\begin{equation}
	\mathcal{E}_{\mathrm{disc}}=\sum_{t}^{t+k}\int_{t_0}^{t_\delta}\left\|\frac{d {z}(t)}{d t}-f({z}(t), t ; \psi)\right\|_2^2 d t,
\end{equation} 
where \(\frac{d {z}(t)}{d t}\) is the actual rate of change of latent states, and \(f({z}(t), t; \psi)\) is the rate of change predicted by NeuralODE. The total loss combines the BC and a discontinuity penalty:
\begin{equation}
\mathcal{L}=\frac{1}{N} \sum^{N}\mathcal{L}_{\mathrm{BC}}+\mathcal{E}_{\mathrm{disc}}.
	\label{allloss}
\end{equation}

\section{Related Work}
\label{related_work}

\noindent\textbf{Language-Conditioned Manipulation (LCM).}
LCM enables robots to perform tasks via natural language, making it a key paradigm in embodied AI by integrating vision, language, and robotics \cite{guhur2023instruction,chen2024embodied}. 
A key approach to tackling LCM tasks is Behavioral Cloning (BC), which trains policies for sequential decision-making directly from expert demonstrations
\cite{
kober2009learning,
foster2024behavior}.
Recent work explores vision-based representations and vision-language interaction to enhance policy learning.
Voxel-based methods like PerAct \cite{shridhar2023perceiver} and large-scale models like RT-1 \cite{brohan2022rt}
enable language grounding but suffer from high computational and parameter costs.
Thus, we propose a lightweight framework that predicts actions from RGB and robot states, while preserving fine-grained linguistic-physical alignment.

\noindent\textbf{Compounding Errors in Behavioral Cloning.} 
BC models an action distribution given the current state at each timestep \cite{
wang2024diffail}. 
However, BC suffers from compounding errors that lead to distributional drift and unrecoverable states over long horizons
\cite{mehta2025stable}. Prior works include interactive corrections
require expert feedback during execution and are impractical for complex teleoperation \cite{hoque2023fleet}, while data augmentation \cite{
lee2015learning,
deshpande2024data} reduces supervision but struggles with fine manipulation.
Temporal abstraction methods \cite{zhaolearning,lukoi} 
discretize actions into phases, but often induce jerky motions.
R3M \cite{nair2023r3m} attempts to align text with vision but lacks dynamic adaptation to evolving task contexts.
In contrast, we jointly optimize stepwise semantic grounding and temporal continuity, mitigating compounding errors and the rigidity in BC.

\noindent\textbf{Neural Ordinary Differential Equations.}
Neural Ordinary Differential Equations (NeuralODEs) extend neural networks to model continuous-time dynamics, evolving from discrete ResNet \cite{he2016deep} updates to continuous trajectories \cite{bilovs2021neural}. 
ODE-RNNs \cite{rubanova2019latent} 
are designed to model
irregularly sampled time series, while TrajODE \cite{liang2021modeling} focuses on enhancing functionalities of recurrent neural networks.
Studies \cite{lin2021no,li2024hides} show that learnable control in NeuralODEs approximates a nonlinear dynamics-inversion controller.
Although NeuralODEs have been utilized in robotic controllers, we harness them for high-level decision-making, leveraging their continuous dynamics to handle the complexities of high-dimensional visual tasks.

\section{Experiments}
\label{experiments}
We conduct comprehensive evaluations against state-of-the-art behavior cloning baselines and ablate key components of our method. We further examine how stepwise semantic anchoring improves visual-action alignment, and how NeuralODE solver parameters affect trajectory continuity.

\subsection{Experimental Setup}
\textbf{Simulation Environments and Datasets.} We evaluate CCoL on three widely used simulation suites: (1) We conduct two bimanual collaborative tasks on Aloha MuJoCo \cite{zhaolearning}, using human and scripted demonstrations under varied object positions. 
(2) We employ RLBench \cite{james2020rlbench} for 
multi-scenario evaluation. (3) Following \cite{zeng2024learning}, we evaluate CCoL on multi-stage Franka Kitchen tasks to assess long-horizon performance.

\noindent\textbf{Baselines.} 
We evaluate CCoL against representative BC methods across three categories. 
\textit{Temporal modeling baselines} include BCCNN \cite{jang2022bc}, RT-1 \cite{brohan2022rt}, BeT \cite{shafiullah2022behavior}, and VINN \cite{pari2022surprising}, mapping states to actions
via CNN, transformers, non-parametric retrieval and weighted k-nearest neighbors. 
ACT \cite{zhaolearning} and AWE \cite{shiwaypoint} reduce planning horizon via action chunking and waypoint abstraction.
\textit{Diffusion-based methods} like DP \cite{chi2023diffusion} and DIC \cite{xu2025diffusion} model actions via conditional denoising diffusion. HDP \cite{ma2024hierarchical} and 3DDiff \cite{ke20243d} leverage 3D scene tokens to decompose policy into joint-level diffusion.
\textit{Representation-enhanced policies} like R3M (Nair et al. 2023), Voltron (Karamcheti et al. 2023), and MPI (Zeng et al. 2024) leverage  visual representations to improve semantic grounding.

\begin{table}[t]
\centering
\setlength{\tabcolsep}{1mm}
\belowrulesep=0.3pt
\begin{tabular}{lcccccc}
\toprule
\multirow{2}{*}{\textbf{Method}} & \multicolumn{3}{c}{{Cube Transfer}} & \multicolumn{3}{c}{{Bimanual Insertion}} \\
\cmidrule(lr){2-4} \cmidrule(lr){5-7}
& Scripted & Human & Avg. & Scripted & Human & Avg. \\
\midrule
BCCNN    & 1.0  & 0.0  & 0.5  & 1.0  & 0.0  & 0.5  \\
RT-1     & 2.0  & 0.0  & 1.0  & 1.0  & 0.0  & 0.5  \\
VINN     & 3.0  & 0.0  & 1.5  & 1.0  & 0.0  & 0.5  \\
BeT      & 27.0 & 1.0  & 14.0 & 3.0  & 0.0  & 1.5  \\
DP      & 54.0 & 4.0 & 29.0 & 74.0 & 0.0 & 37.0 \\
ACT      & 86.0 & 50.0 & 68.0 & 32.0 & 20.0 & 26.0 \\
AWE      & \underline{99.0} & 71.0 & 85.0 & 57.0 & 30.0 & 43.5 \\
DIC      & 95.9 & \underline{78.1} & \underline{87.0} & \underline{83.2} & \underline{30.2} & \underline{56.7} \\
\textbf{CCoL} & \textbf{99.0} & \textbf{82.0} & \textbf{90.5} & \textbf{87.0} & \textbf{36.0} & \textbf{61.5} \\
\bottomrule
\end{tabular}
\caption{\normalsize  Success rate (\%) on Aloha MuJoCo.
}
\label{table1}
\end{table}

\begin{table}[t]
\centering
\setlength{\tabcolsep}{1mm}
\belowrulesep=0.3pt
\begin{tabular}{lccccc}
\toprule
\textbf{Method} & LampOn & GrillMeat & Phone& OpenBottle & Avg. \\
\midrule
BCCNN   & 4.0  & 0.0  & 0.0  & 0.0  & 1.0  \\
ACT     & 76.0 & 70.3 & 22.0 & 42.7 & 52.8 \\
AWE     & \underline{85.7} & \underline{74.3} & \underline{34.7} & \underline{46.3} & \underline{60.3} \\
\textbf{CCoL}    & \textbf{93.7} & \textbf{82.3} & \textbf{44.3} & \textbf{51.7} & \textbf{68.0} \\
\midrule
HDP     & \underline{92.7} & 80.7 & 42.3 & 50.7 & 66.6 \\
3DDiff   & 89.3   & \underline{85.0}   & \underline{71.7}   & \underline{69.3}   & \underline{78.8}   \\
\textbf{CCoL$_\text{3D}$} & \textbf{97.3} & \textbf{87.3} & \textbf{76.7} & \textbf{78.3} & \textbf{84.9} \\
\bottomrule
\end{tabular}
\caption{Comparison on RLBench tasks (3 seeds). 
}
\label{table2}
\end{table}

\begin{table*}[t]
\centering
\setlength{\tabcolsep}{1.9mm}
\belowrulesep=0.3pt
\begin{tabular}{llcccccccc}
\toprule
\multirow{2}{*}{\textbf{Method}} & \multirow{2}{*}{\textbf{Backbone}} 
& \multicolumn{4}{c}{\textbf{Single-task}} 
& \multicolumn{4}{c}{\textbf{Long-horizon task}} \\
\cmidrule(lr){3-6} \cmidrule(lr){7-10}
& & \textcircled{1} Turn Knob & \textcircled{2} Open Door & \textcircled{3} Open Microwave & Avg. 
 & \textcircled{1}+\textcircled{2} & \textcircled{2}+\textcircled{3} & \textcircled{1}+\textcircled{2}+\textcircled{3} & Avg. \\
\midrule
INSUP      & ResNet50  & 28.0 & 18.0 & 26.7 & 24.2 & --   & --   & --   & --   \\
CLIP       & ResNet50  & 26.3 & 13.0 & 24.7 & 21.3 & --   & --   & --   & --   \\
R3M        & ResNet50  & 53.3 & \underline{50.7} & 59.3 & 54.4 & 25.1 & 29.6 & 15.5 & 23.4 \\
Voltron    & ViT-S     & 71.7 & 45.3 & 40.3 & 52.4 & 34.6 & 22.6 & 13.0 & 23.4 \\
MPI        & ViT-S     & \underline{83.3} & {50.3} & \underline{59.7} & \underline{64.4} & \underline{38.2} & \underline{31.2} & \underline{23.3} & \underline{30.9} \\
\textbf{CCoL} & ViT-S & \textbf{85.3} & \textbf{56.7} & \textbf{64.7} & \textbf{68.9} & \textbf{43.7} & \textbf{37.3} & \textbf{27.7} & \textbf{36.2} \\
\midrule
MPI        & ViT-B     & \textbf{89.0} & \underline{57.7} & {54.0} & \underline{66.9} & \underline{41.7} & {34.7} & {26.3} & {34.2} \\
\textbf{CCoL} & ViT-B$_\text{frozen}$  & {81.7} & {53.3} & \textbf{62.7} & {65.9} & {41.6} & \underline{35.4} & \underline{26.5} & \underline{34.5} \\
\textbf{CCoL} & ViT-B  & \underline{88.7} & \textbf{58.3} & \underline{58.7} & \textbf{68.6} & \textbf{45.7} & \textbf{38.5} & \textbf{30.1} & \textbf{38.1} \\
\bottomrule
\end{tabular}
\caption{\normalsize Performance comparison on Franka Kitchen. \textbf{Bold} and \underline{underlined} values mark the best and second-best performance.
}
\label{kitchen}
\end{table*}

\noindent\textbf{Metrics and Implementation Details.}
The success rate (\%) measures task completion under defined conditions, such as transferring a red cube between grippers with 1cm clearance to avoid collisions.
To ensure consistency, our model maintains the same settings as AWE \cite{shiwaypoint,ke20243d} for Aloha MuJoCo and RLBench, and aligns with MPI \cite{zeng2024learning} for the Franka Kitchen setup. The model employs the SGD optimizer, starting with a learning rate of 1e-5, momentum of 0.9, a chunking size of $k=$50, and a batch size of $8$. 
The ODEsolver evaluates solutions at two discrete time points.

\begin{table}[t]
\centering
\setlength{\tabcolsep}{1mm}
\belowrulesep=0.3pt
\begin{tabular}{lcccc}
\toprule
\multirow{2}{*}{\textbf{Method}} & \multicolumn{2}{c}{{Cube Transfer}} & \multicolumn{2}{c}{{Bimanual Insertion}} \\
\cmidrule(lr){2-3} \cmidrule(lr){4-5}
& Scripted & Human & Scripted & Human \\
\midrule
\textbf{CCoL} & \textbf{99.0} & \textbf{82.0} & \textbf{87.0} & \textbf{36.0} \\
\quad \(\lrcorner\) w/o MCC & 99.0 & 74.0 & 72.0 & 34.0 \\
\quad \(\lrcorner\) w/o CSA & 99.0 & 73.0 & 74.0 & 35.0 \\
\quad \(\lrcorner\) w/o $\mathcal{E}_{\mathrm{disc}}$ & 99.0 & 78.0 & 76.0 & 32.0 \\
\quad \(\lrcorner\) w/ CSA$_\text{no att}$ & 98.0 & 72.0 & 70.0 & 28.0 \\
\quad \(\lrcorner\) w/ TCN & 96.0 & 72.0 & 74.0 & 36.0 \\
\quad \(\lrcorner\) w/ TCN$_{\text{no MCC}}$ & 92.0 & 58.0 & 54.0 & 20.0 \\
\bottomrule
\end{tabular}
\caption{\normalsize Ablation study of MCC and CSA variants.}
\label{table_abl}
\end{table}

\begin{figure}[t]
	\centering 
    \includegraphics[height=0.2\textwidth]
    {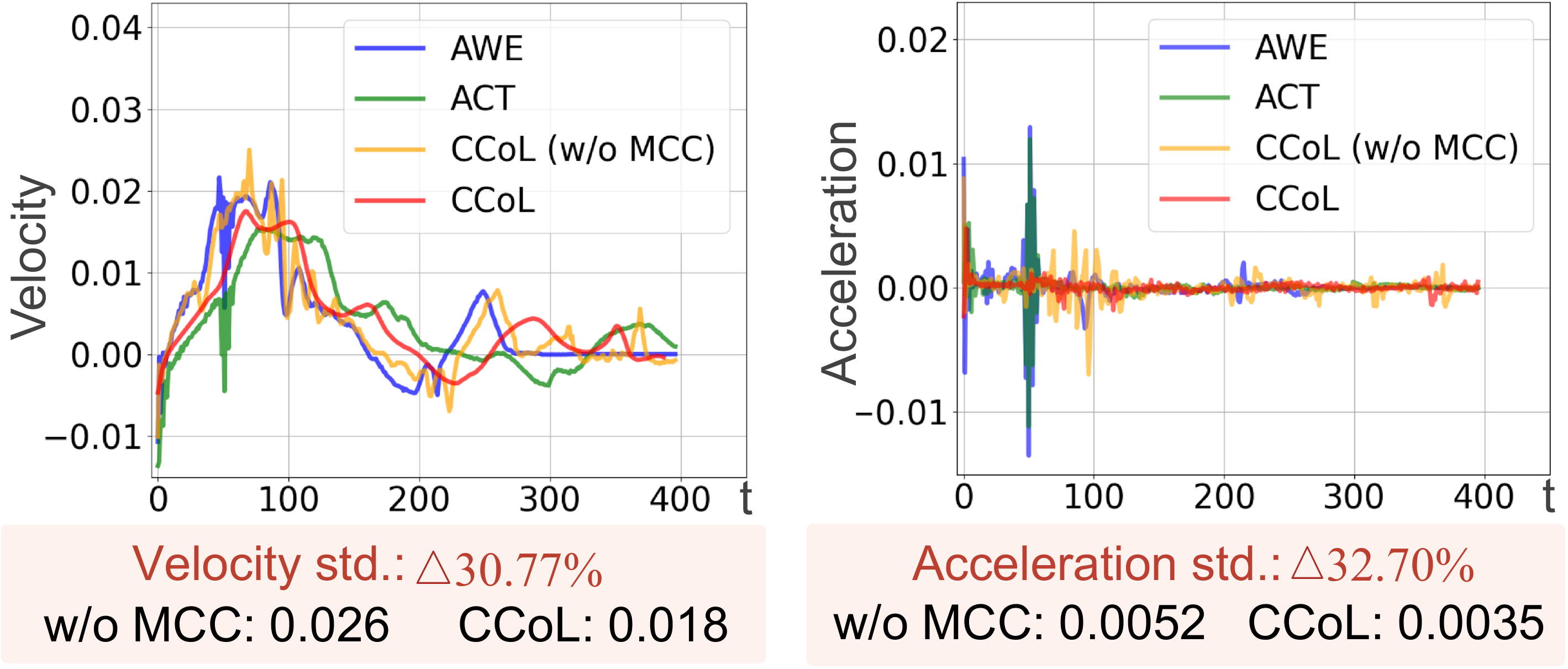} 
\caption{Trajectory smoothness analysis for the shoulder joint during the bimanual insertion scripted task.
}
	\label{smooth1}
\end{figure}
\subsection{Experiment Results}
\textbf{(1)} As shown in Table~\ref{table1}, CCoL excels in bimanual coordination tasks, surpassing AWE by +11.8\% (absolute) and DIC by +5.8\% (relative) in average success rate. In particular, it achieves +19.2\% relative gain over DIC in human-demonstrated bimanual insertion, showing robustness to noisy supervision and task complexity.
\textbf{(2)} On RLBench (Table~\ref{table2}), CCoL surpasses AWE by +7.7\% in 2D setting, and CCoL$_{\text{3D}}$ outperforms 3DDiff by +6.1\%, highlighting the benefit of multimodal coordination in 3D spatial domains.
For a fair comparison, CCoL$_{\text{3D}}$ aligns its 3D representation resolution with 3DDiff by leveraging RGB-D input, a CLIP-based ResNet-50 encoder, and shared 3D tokens with relative attention. 
\textbf{(3)} We evaluate CCoL with multiple vision backbones (Table~\ref{kitchen}). Under ViT-S (22M), it improves over MPI by +6.9\% and +17.2\% on single-task and long-horizon settings, respectively. With ViT-B (86M), it achieves +11.4\% improvement on the long-horizon setting, and notably reaches 34.5\% even with a frozen encoder, showing strong performance without vision tuning.

\subsection{Ablation Study}
\noindent\textbf{Quantitative Analysis.}
Table \ref{table_abl} reveals that w/o (i.e., without) MCC causes a 15.0\% drop on bimanual insertion (scripted) due to disrupted continuity through latent proprioceptive modeling, while w/o CSA leads to a 9.0\% average drop on cube transfer (human), 
highlighting its importance in aligning linguistic with visuomotor context. 
Removing $\mathcal{E}_{\mathrm{disc}}$ from MCC degrades performance by weakening temporal smoothness. CSA$_{\text{no att}}$ replacing bidirectional attention with average pooling performed worse than completely w/o CSA. Replacing attention with TCN in the proprioceptive encoder reduces success by 13.0\% in the bimanual insertion (scripted) task, while further removing MCC leads to a 16.0\% drop on human demos, showing that both temporal modeling and attention are essential for coherent behavior. 


\begin{figure*}[t]
    \centering
    \includegraphics[width=0.97\linewidth]
    {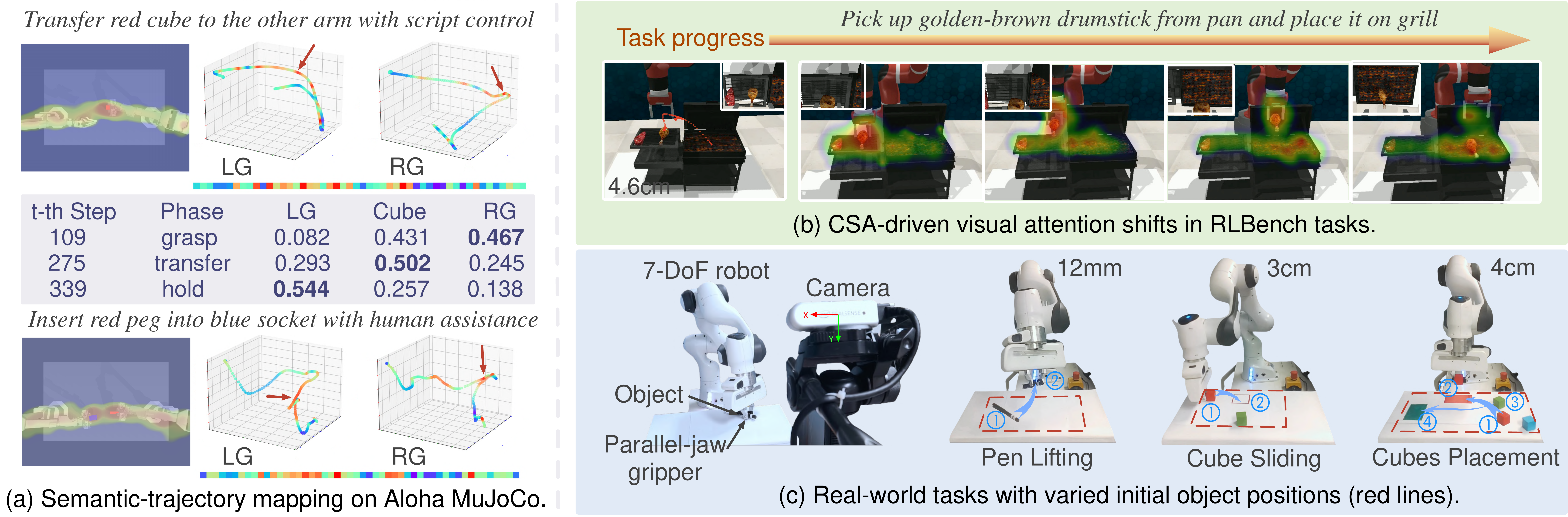}
    \caption{Attentive attribution mapping via CSA and real-world task settings with varied initial object positions.
    }
    \label{fig:heatmap2}
\end{figure*}

\begin{figure}[t]
	\centering 
    \includegraphics[height=0.19\textwidth]
    {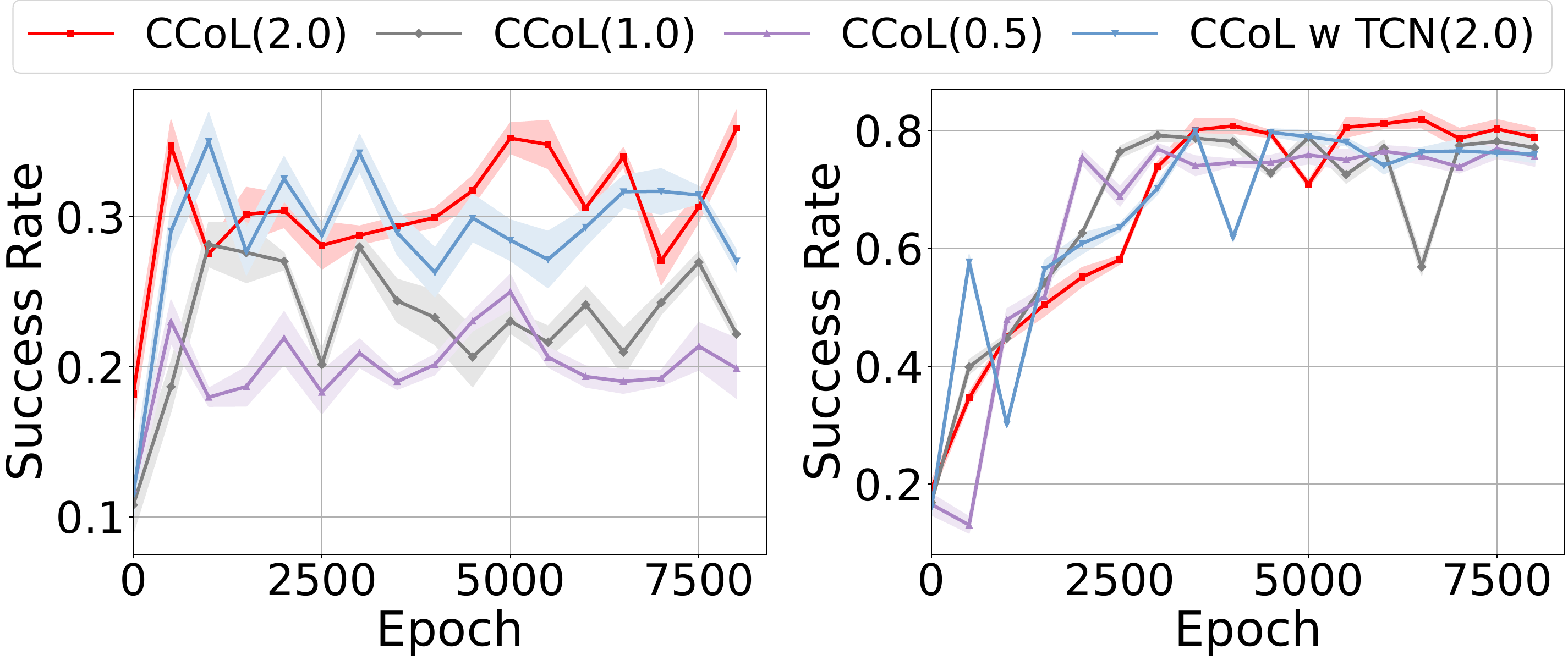} 
\caption{Hyperparameter impact on insertion and transfer.
}
	\label{hyper}
\end{figure}
\noindent\textbf{Trajectory Smoothness Analysis.} To assess CCoL's physical continuity, we compute velocity $v(t)= \frac{d a^{\prime}_{t}}{d t} \approx a^{\prime}_{t+1}-a^{\prime}_{t}$ and acceleration $c(t)= \frac{d^2 a^{\prime}_{t}}{d t^2} \approx a^{\prime}_{t+2}-2 a^{\prime}_{t+1}+a^{\prime}_{t}.$ 
As shown in Fig. \ref{smooth1}, CCoL 
reduces high-frequency oscillations (e.g., peak acceleration) compared to baselines.
This smoothness stems from MCC’s latent proprioceptive modeling via NeuralODEs (Eq. \ref{eqode}), which enables continuous transitions. Compared to CCoL$_{\text{w/o MCC}}$, CCoL reduces velocity fluctuations by 30.8\% and acceleration fluctuations by 32.7\%. Remarkably, the minimum acceleration improved 20.2\%. These findings validate the ability of MCC to suppress high-frequency jitter and severe negative acceleration (sudden deceleration), resulting in smoother transitions.

\noindent\textbf{Hyperparameter Analysis.} Fig. \ref{hyper} shows the effect of NeuralODE solver timesteps (2.0, 1.0, 0.5) on long-horizon task performance. Larger timesteps yield higher success rates and faster convergence, especially in cube transfer (scripted), 
supporting our design that coarse steps improve efficiency while preserving smooth dynamics. Notably, CCoL(2.0) shows stable monotonic improvement, whereas smaller steps (e.g., 0.5) 
lead to oscillations from sensitivity to transient noise.

\noindent\textbf{Qualitative Analysis.} 
We analyze how linguistic cues influence visual attention and trajectory planning through CSA. 
Fig.~\ref{fig:heatmap2}(a) illustrates that CSA aligns nouns (e.g., “cube”, “socket”) to visual targets while verbs (e.g., “transfer”, “insert”) map to distinct trajectory patterns. 
We also evaluate CSA’s temporal grounding behavior by analyzing attention shifts across tasks. In cube transfer, quantitative attention scores show a clear progression from the right gripper (grasp) to the red cube (transfer) and then the left gripper (handover). A similar shift is observed in the RLBench grillmeat task (Fig. \ref{fig:heatmap2}(b)).
This phase-to-phase transition shows that CSA dynamically adjusts attention based on semantic cues, enabling step-aware visuomotor alignment.

\begin{figure}[t]
	\centering 
    \includegraphics[height=0.23\textwidth]
    {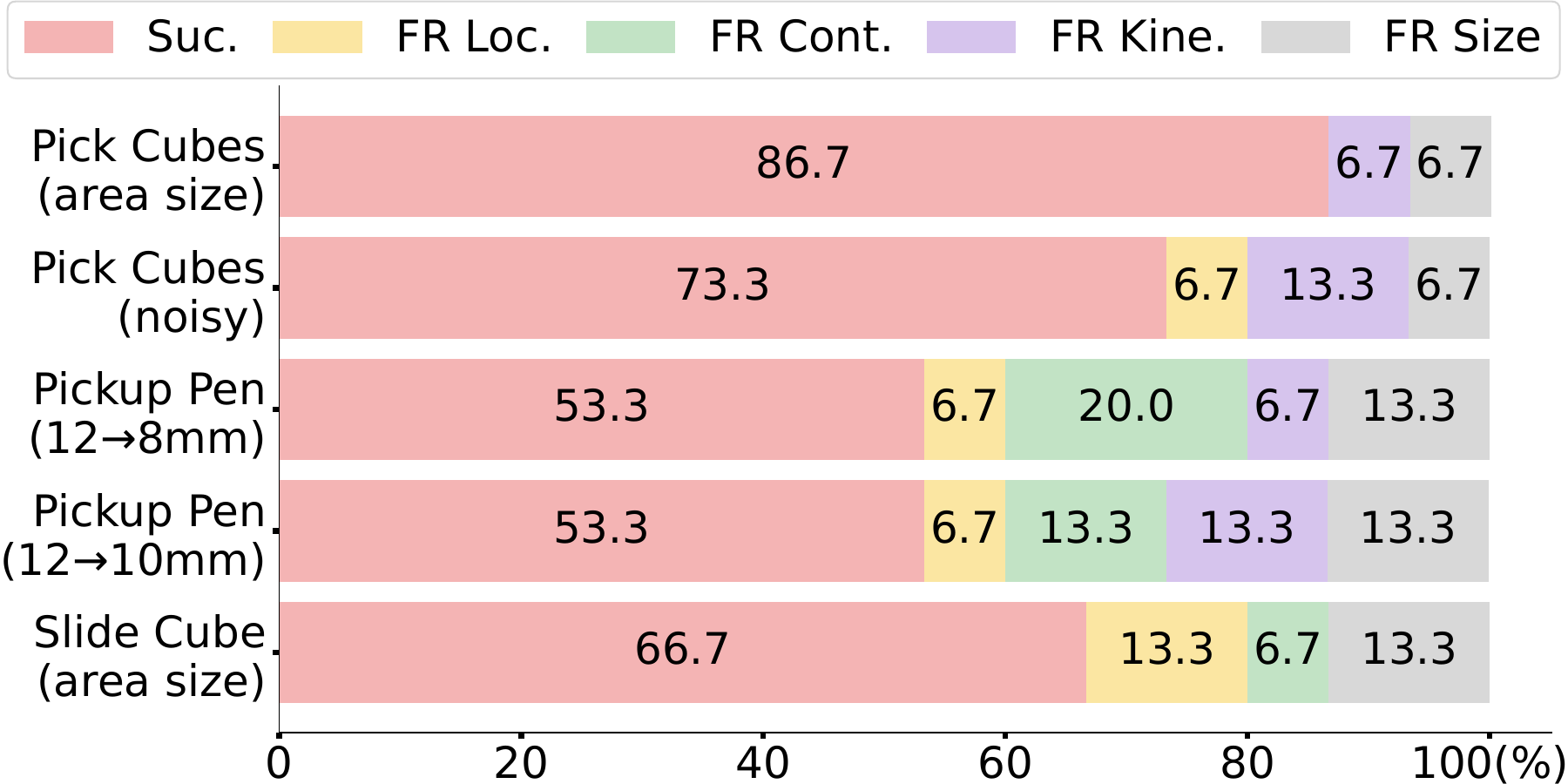} 
\caption{Real-world generalization under unseen states. Bars indicate success and four types of failure rates.
}
	\label{real}
\end{figure}

\noindent\textbf{Real-World Experiments.} 
To validate our framework in practical scenarios, we deploy it on a 7-DoF Franka Emika Panda robot equipped with an Intel RealSense D435i RGB-D camera and a parallel-jaw gripper (Fig. \ref{fig:heatmap2}(c)). We design three manipulation tasks (pen lifting, cube sliding and cubes placement) to assess generalization under unseen (e.g., varying pen diameters) or noisy states (e.g., vase, pen holder).  
For training, we collect 50 kinesthetic demonstrations per task with randomized initial conditions. At test time, each task is evaluated over 15 trials under shifted conditions.
As shown in Fig. \ref{real}, CCoL consistently achieves high success rates (e.g., 86.7\% in cubes placement), demonstrating strong real-world generalization.
Since failures are mainly caused by localization errors, contact (grasping) failures, kinematic inconsistencies, and size adaptation issues, we report the average failure rate (FR) of each failure.
We found that when picking cubes, placement errors occurred in small color areas, despite successful localization and contact. When picking up a pen, failures were caused by small diameters and insufficient lifting distances. When sliding a cube, failures stem from positioning deviations or incomplete sliding into the target area. Our model trains in 5.3 hours on an RTX 4090 GPU and runs at 0.015s (±0.003s) per action sequence, meeting real-time 
requirements ($\approx 67 \text{Hz}$ policy frequency). 


\section{Conclusion}
\label{conclusions}
We present CCoL, a novel framework that synergizes multimodal continuous co-learning and stepwise semantic-physical alignment to mitigate compounding errors in BC. By integrating NeuralODEs for smooth latent trajectory learning and bidirectional cross-attention for linguistic grounding, CCoL ensures both temporal coherence and semantic precision during task execution.
Extensive simulation experiments, alongside real-world robustness, demonstrate its superiority, especially in bimanual collaborative and long-horizon manipulation tasks. Future work will extend CCoL to LLM-based methods and explore its integration with foundation models for open-world manipulation.

\section{Acknowledgments}
This work was supported by National High Level Hospital Clinical Research Funding (Grant 2025-PUMCH-D-005), National Natural Science Foundation of China (Grant 62473213), Sustainable Development Science and Technology Special Project of Shenzhen (Grant KCXFZ20230731100900002), Shenzhen Science and Technology Program (Grant KQTD20210811090143060), and Beijing Tianjin Hebei Basic Research Cooperation Special Project (Grant 24JCZXJC00060). 
We also acknowledge support
in part 
from HK RGC General Research Fund (No.: PolyU 15235424), ``Research on Key Technologies for Systematic Artificial Intelligence Agents'' project under China Mobile Innovation and Research Institute (No.: R24114H7), and Research Institute for Artificial Intelligence of Things, The Hong Kong Polytechnic University.
This work was partially conducted at the Robotics and Embodied Intelligence (REI) Lab, The Education University of Hong Kong.

\bibliography{bib/aaai2026}

\end{document}